\documentclass[twocolumn,10pt]{IEEEtran}
\def\BibTeX{{\rm B\kern-.05em{\sc i\kern-.025em b}\kern-.08em
T\kernf-.1667em\lower.7ex\hbox{E}\kern-.125emX}}

\begin{document}

\vspace{-0.5cm}
\title{Joint Estimation of Information and Distributed Link-Scheduling in Wireless Networks: Mean-Field Approximation and Graphical Models}

{\small
\author{Sung-eok Jeon, and Chuanyi Ji\\
sujeon@microsoft.com, jic@ece.gatech.edu
}
}

\maketitle

%%%%%%%%%%%%%%%%%%

\begin{abstract}
In a large multi-hop wireless network,  nodes are preferable to make distributed link-scheduling decisions with information exchange only among a small number of neighbors. However, for a slowly-decaying channel  and densely-populated interfering nodes, a small  size neighborhood  often  results in nontrivial link outages and is thus insufficient for making optimal scheduling decisions. A question arises how to incorporate the information outside a neighborhood in distributed link-scheduling. In this work, we develop joint approximation of information and distributed link scheduling.   We  first apply machine learning approaches to model  distributed link-scheduling  with complete information. We then characterize the information outside a neighborhood, i.e., the residual interference, as an aggregated random loss variable. The loss variable is characterized by either a Mean-Field approximation or a normal distribution based on the Lyapunov central limit theorem. The approximated information outside a neighborhood is incorporated in a factor graph. This results in  joint approximation of information and distributed link-scheduling in an iterative fashion. Link-scheduling decisions are first made at each individual node based on the approximated  loss variables. Loss variables are then updated and used for  next  link-scheduling decisions. The algorithm repeats between these two phases until convergence. Interactive iterations among these variables are implemented by a message-passing algorithm over a factor graph. Simulation results show that using learned information outside a neighborhood jointly with distributed link-scheduling reduces the outage probability close to zero even for a small neighborhood. 
\end{abstract}


%%%%%%%%%%%%%%%%%%
\section{Introduction}
\label{Introduction}

Efficient link-scheduling and channel assignment are fundamental  to optimal resource utilization in wireless networks \cite{Chiang}\cite{Han}\cite{Modiano}. 
Distributed scheduling is a frequent scenario, where individual nodes make scheduling decisions using only information from their neighbors \cite{Bjorklund}\cite{MengJi}\cite{Modiano}\cite{Tan}. 

Such distributed scheduling is optimal if an objective function to be optimized is local by nature, i.e., decomposable to factors with nodal variables and their neighbors \cite{MengJi}\cite{Modiano}\cite{Tan}. Distributed scheduling with local information-exchange becomes non-optimal if information outside neighborhoods is pertinent. For example, interference causes long-range spatial dependence in multi-hop wireless networks \cite{IEEESigProc}\cite{Madan}; thus optimal scheduling decisions made by nodes require information not only within neighborhoods but also outside. 

In our prior work, we studied such a scenario where there was a long range spatial dependence among nodes but the dependence was weak. We developed distributed scheduling based on a machine learning framework, where information exchange occurred only among neighbors. Information, i.e., on residual interference, was entirely ignored outside a neighborhood range. We derived a  sufficient condition to show that the aggregated spatial dependence outside a neighborhood needed to be weak, i.e., decay sufficiently fast, for such a distributed algorithm to be near-optimal. 

This work studies the scenarios that the spatial dependence is of long-range and strong. The information, if ignored outside neighborhoods, would result in a significant link outage probability. Therefore, distributed scheduling with information exchange only from neighbors is no longer 
near-optimal\cite{IEEESigProc}. This corresponds to the scenario of either a slowly-decaying channel or densely populated interferers. There, the aggregated residual interference cannot be ignored \cite{Andrews}\cite{Pico}\cite{Elbatt}\cite{Madan}\cite{IEEESigProc}. Hence, a node needs to possess information outside its neighborhood, e.g., on residual interference, when making cooperative scheduling-decisions. Otherwise, decisions on channel-access made by a node may cause significant interference even at faraway nodes in a network. In our prior work \cite{IEEESigProc}, we show that residual interference outside a neighborhood cannot be ignored especially for slowly decaying channels and densely populated nodes. However, how to incorporate the information outside a neighborhood is not studied there. Overall, systematic study is also lacking on what size neighborhood is sufficient and what information to include for optimal distributed link-scheduling.

 One scenario  is for nodes to simply include complete information outside their neighborhood. For example, a protocol model based on contention  \cite{Shi} assumes that each node maintains the configuration information from an entire network. This, however, does not reflect a realistic communication environment \cite{IEEESigProc}, where complete information is often unavailable at individual nodes. Even if the complete information were available, a node would need to exchange information with all others \cite{Elbatt}\cite{Gandham}\cite{Lu}, which can be  prohibitive in a large network. As an alternative, we consider aggregated information outside the neighborhoods as an approximation. 

The approximation  does not consider information from individual nodes outside a neighborhood but their aggregated contribution. Such approximation needs to satisfy a certain optimality in regard to an underlying exact model. Hence, we first provide a network model for distributed link-scheduling. The model is developed in our prior work \cite{IEEESigProc}, and based on Boltzmann distribution and probabilistic graphical models (see \cite{Geman} \cite{Jordan} and reference therein). The model includes long-range spatial dependence among all nodes in a network, and is exact under given assumptions.
 
We then consider two approximations to the long-range dependence, i.e., residual interference, outside a neighborhood range. Both approximations regard the  information outside a neighborhood as one aggregated variable.  In particular, the first approximation views the aggregated residual interference as a  deterministic mean-field.  The mean-field parameters optimize a variational bound of the probabilistic graph of our network model \cite{MacKay}. The parameter values are obtained through solving coupled mean-field equations \cite{Jordan}\cite{MacKay}. 
Mean-field approximations  have been used widely in machine learning and image processing \cite{Jordan}  in the context of graphical models. In networking, Mean-Field approximation has been applied to loss networks \cite{Kelley}. These forms of Mean-Field approximations are derived based on an underlying model of networks, which motivate our work here. For wireless network, Mean-Field has been considered  as a limiting form of a large number of mobile users 
\cite{Chaintreau}. This is related to the idea of aggregating dependencies outside a neighborhood in this work. Mean-Field theory is also used  to derive effective link capacity of wireless networks \cite{Marbukh}.  

The second approximation  treats the aggregated residual interference as a random variable, referred to as loss variable. We show that the loss variable can be approximated  by a normal distribution through the  Lyapunov central limit theorem (CLT) \cite{Stat}.  The mean and the variance of the normal distribution can be learned using information within a neighborhood range. 
The approximations are combined with iterative and distributed link-scheduling decisions. In particular, a message-passing algorithm on the factor graph \cite{Kschischang} is combined with the approximations of the long-range residual interference for the distributed and statistical link-scheduling. The mean-field parameters and the normal residual random variable form  particular local functions in the factor graph.  The Mean Field approximation or the distribution of the loss variables are learned iteratively from the available information within the neighborhood range. Distributed decisions are then made through alternating between approximation and message passing over the factor graph. This process repeats until  convergence. 

We then conduct simulation on network-wide link outage probability to complement the analytical derivation of the approximation and the algorithm. Simulation results show that distributed link-scheduling with an arbitrarily small neighborhood and aggregated information outside can satisfy the global SINR requirements while maximizing the channel reuse.

The rest of the paper is composed as follows.  Section II provides motivating examples on distributed scheduling using complete, partial and approximated information. In Section \ref{PF}, we provide a problem formulation.  Section \ref{WNModel} provides  an accurate  probabilistic model of distributed link-scheduling decisions at individual nodes. 
In Section \ref{Graph3}, we obtain two approximations on residual interference, and the resulting simplified probabilistic graphical models. 
In Section \ref{SelfManagement}, we derive distributed message-passing algorithm coupled with the approximation. In Section \ref{PE}, we provide simulations to show the performance of the algorithms using two approximations. Section \ref{Conclusions} concludes this study and discusses open questions.

\section{Distributed Link-Scheduling with Loss Variables}

In this section, we provide motivation on the importance of information outside a neighborhood. We first describe the setting of distributed link-scheduling. We then present an example of distributed scheduling with complete information. We  provide an example how the performance degrades for distributed scheduling with partial information. This motivates two approximation schemes. We provide an overview on the approximation schemes to motivate the subsequent sections.  

\subsection{Distributed Link-Scheduling Decisions}

In a distributed setting, a node determines its channel access schedule, i.e., whether to access a channel at a time slot, using the information within its neighborhood. The information is on the ``state" that includes the relative positions, the transmission statues (who is transmitting to whom), transmission power, and scheduling decisions of links. For simplicity, this work does not consider power control and  assumes fixed transmission power for each node. 

The decision made is cooperative and based on two criteria. The first is whether the SINR requirements are satisfied at the neighboring links if a node chooses to use a channel. The second is to maximize the spatial channel-reuse which is the number of concurrently active links in the network. 

Decisions are made in a distributed fashion. A node sends its decision to neighbors. The neighbors use the information to make their own link-scheduling decisions, and then send the decisions to their neighbors. Distributed decisions are made iteratively and collectively by all nodes in a network until convergence.

An assumption is that a node knows complete information from its neighbors. Hence, the neighborhood size relative to the entire network determines whether the information used by a node is complete or partial.

\subsection{Complete Information}

First, we consider an ideal setting, where complete information is available for individual nodes to make link-scheduling decisions \cite{Ergen}\cite{Rhee}. The complete information is on the state from all nodes in a network. That is, the neighborhood for any node is an entire network. A node then determines whether to access a channel at a timeslot using the available information. Such a scenario is used as a baseline for comparing approximations.
 
\begin{figure}[htb!]
\epsfysize=0.4in
\centerline{\epsffile{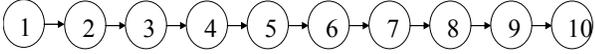}}
\caption{A Linear Topology with $10$ Nodes and $9$ Directional Links.}
\label{Linear_10}
\end{figure}

Consider an example of a linear network in Figure \ref{Linear_10}. Assume that the channel-contention constraint requires any active link between two adjacent nodes to be separated by one neighboring link. The interference constraint requires any active link to be separated by at least two links. Assume also that the network achieves the spatial channel-reuse maximization where the total number of concurrent active links is maximized. 

\indent
Figure \ref{Feasibles} shows all possible configurations of active links. Each row corresponds to a snapshot of active links at a time epoch, where the active links  satisfy the aforementioned contention and interference constraints. For example, the first row denotes a configuration where both link ($3,4$) and ($8,9$) are active. Overall, there are $2$ configurations with two active links.  The remaining $8$ configurations have three active links that result in the maximum channel reuse.  Note that for distributed decisions where each node has complete information on the activities of the others, the configurations with three active links are ones that maximize the channel reuse while satisfying the constraints. 

\subsection{Partial Information}

Partial (state) information is available to a node when its neighborhood  consists of only nodes from parts of a network.

\begin{figure}[htb!]
\epsfysize=2.0in
\centerline{\epsffile{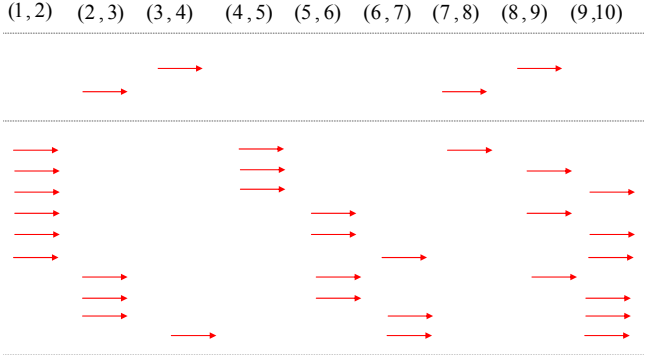}} %Feasibles.eps
\caption{Possible configurations.  ``$\rightarrow$": Active link}
\label{Feasibles}
\end{figure}

For example, assume the neighborhood for each node is the adjacent link. This means that the neighborhood is only determined by the contention constraint where a node would not access the channel if its neighbor is transmitting. Such a neighborhood size is smaller than desirable since node does not know whether SINR is satisfied at its neighbors' neighbors. Hence, the SINR constraint may be violated.

Specifically, under the above assumption on the neighborhood size, it is legitimate for node $3$ to transmit to node $4$ when links $(1,2)$, $(6,7)$ and $(9,10)$ are active (see the eighth row in Figure \ref{Feasibles}). Those active links are outside the neighborhood system and thus invisible to node $2$. Active link $(3,4)$ thus generates interference, violating the SINR constraints at the existing active links. 

By increasing the neighborhood size, we show, in our prior work \cite{IEEESigProc}, that an optimal configuration with three active links can be obtained through fully distributed node decisions with a chosen neighborhood size. For example, if two links $(2,3)$ and $(7,8)$ are active at the beginning, a configuration with $3$ active links can be obtained if the distributed algorithms are randomized, i.e., nodes make probabilistic decisions for transmission. Our prior work also derives sufficient conditions for such distributed scheduling to be near-optimal. However, our prior work does not study sufficiently when partial information in distributed scheduling is insufficient for an obtaining optimal configuration.

%%%%%%%%%%%%%%%%%%%%%%%%%%%%%%%%%%%%%%%
\subsection{Importance of Information Outside Neighborhood}
\label{ResidualInteference}

The above example shows that for a finite and often small number of neighbors, the SINR requirement is not guaranteed \cite{IEEESigProc}\cite{Madan}. In other words, information within a neighborhood can be insufficient for making correct distributed decisions when the information outside the neighborhood is neglected.   

To further illustrate the importance of information outside a neighborhood,  we consider another example of  a multi-hop wireless network. The network has nodes located in a square area of 10 x 10 square meters,  $L$ = 200 links,  $\alpha=4$ as the channel attenuation factor, and SINR threshold $\mbox{SINR}_{th}=10$. 
Here $\alpha=3$ corresponds to slow power decay of the channel, resulting in strong interference and thus dependence among nodes. Such a scenario is not studied in our prior work \cite{IEEESigProc}, where the strong dependence violates the sufficient condition we derive for distributed scheduling with a moderate neighborhood to be near-optimal.

\begin{figure}[htb!]
\epsfysize=3.0in
\centerline{\epsffile{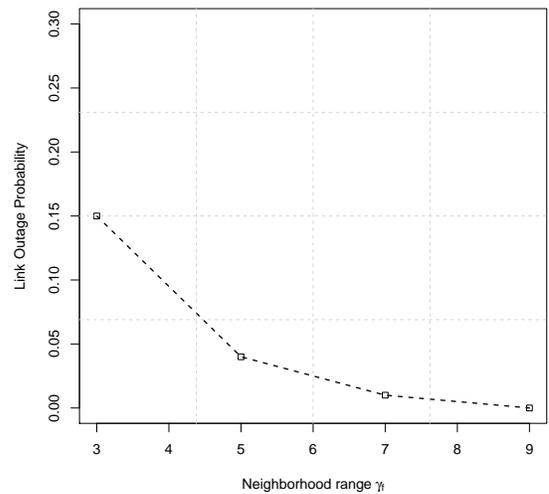}}
\caption{Link Outage probability.}
\label{MotiveWrtRf}
\end{figure}

 Suppose we  ignore the residual interference outside a neighborhood range, which is  common practice for large wireless networks. We then calculate the link outage probability that the SINR requirement is violated at a link. The outage probability  is shown to be significant and not negligible as  in Figure \ref{MotiveWrtRf}, especially for a moderate neighborhood size. 

Note that the link outage due to the ignored residual interference is different from the transient outage due to the fading channel. The link outage here is persistent. Thus, the network connectivity may not be guaranteed. Some nodes can even be permanently disconnected if outages occur persistently at adjacent links.

This example clearly shows that the residual interference cannot be arbitrarily ignored in a large wireless network.

%%%%%%%%%%%%%%%%%%%%%%%%%%%%%%%%%%%%%
\subsection{Approximation of Information Outside Neighborhood}

Now consider that each node accounts for information outside its neighborhood, i.e., the residual interference, as a loss variable. The loss variable treats the impact from the rest of a network except the neighbors in aggregation. 

Two approaches are used for the approximation. The first is to characterize the expected value of a loss variable through the deterministic Mean-Field approximation \cite{Kappen}. That is, a set of Mean-Field equations are obtained through optimizing a variational bound from the network model \cite{MacKay}. The expected values of the loss variables can be obtained iteratively through solving deterministic Mean-Field equations. The second is to regard the loss variable as random and estimate its   probability distribution.  The probability distribution can be learned iteratively using information within the neighborhood. 

The approximations are incorporated into distributed scheduling decisions. Each  node first estimates either the Mean-Field value or the distribution of the loss variable. The node then determines  channel-access schedule using both complete information within its neighborhood and estimated information on the loss variable. The algorithm is to be described in Section \ref{SelfManagement}.

%%%%%%%%%%%%%%%%%%
\section{Problem Formulation}
\label{PF}

We now obtain a problem formulation for distributed scheduling with loss variables. 

%%%%%%%%%%%%%%%%%%%%
\subsection{Notations}

Consider a wireless multi-hop network with $n$ nodes.
$X_i$ and $P_i$ denote the position and the transmission power of node $i$, respectively. 
Let $\mbox{\boldmath$X$}$ = $\{ X_1, \cdots, X_n \}$, where bold characters are used to denote vectors in the rest of the paper. Let $\mbox{\boldmath$L$}$ denote a set of communication links that access a common wireless channel. Let $\sigma_{ij}$ represent the activity of link ($i,j$) $\in$ $\mbox{\boldmath$L$}$ at a time instance:
$\sigma_{ij}$=1 if link ($i,j$) is active, i.e., using the common wireless channel; $\sigma_{ij}$=0, otherwise. Let 
$\mbox{\boldmath$\sigma$}$ = $\{ \sigma_{ij}\}$ where ($i,j$) $\in$ $\mbox{\boldmath$L$}$. Here node $i$ is assumed to be a transmitter whereas node $j$ is a receiver.

We assume that each node maintains complete information  within a neighborhood range. The shape of the neighborhood for each node can be arbitrary rather than  circular. 
The neighborhood range can be different  for all nodes but  bounded by a radius $\gamma_f$. Hence, without loss of generality, $\gamma_f$ is assumed to be the neighborhood range for each node. 

The channel gain for link ($i,j$) between node $i$ and $j$ is denoted as
$G(i,j)$. For simplicity,  $G(i,j) = |X_i-X_j|^{-\alpha}$, where $\alpha$ is the channel power attenuation factor and $2 \le \alpha \le 6$. $G(i,j)$ is random when positions $X_i$'s are assumed random. 

$\mbox{R}_{ij}$ is the inverse of SINR of an active link ($i,j$) at a time instance, 
\begin{equation}
\label{R_gamma}
\mbox{R}_{ij} = { \sum_{mn \neq ij} P_m G(m,j) \sigma_{mn} + \eta_j  \over P_i G(i,j)},
\end{equation}
\noindent
where $\eta_j$ denotes the noise power at the receiver $j$. For simplicity, all nodes  assume to have the same (random) noise power $\eta_j=N_b$. To satisfy the SINR constraint, $\mbox{R}_{ij} \le \mbox{R}_{th}$ holds, where $\mbox{R}_{th}$ = $1/\mbox{SINR}_{th}$ is the inverse of a required SINR threshold $\mbox{SINR}_{th}$.
Let $\mbox{R}^l_{ij}$ be the approximation of $\mbox{R}_{ij}$ that includes complete information within the neighborhood of node $i$ plus the residual interference outside,
\begin{equation}
\label{R_l_gamma}
\mbox{R}^l_{ij} = { \sum_{|X_m-X_j| \le \gamma_f} P_m G(m,j) \sigma_{mn} + \eta_j + \mbox{Res}_{ij} \over P_i
G(i,j)},
\end{equation}
\noindent
where $\mbox{Res}_{ij}$ denotes the approximated value of residual interference outside the neighborhood range experienced by the receiver $j$ of active link ($i,j$). The distance between the closest interferer and the receiver of the active link is referred to as the contention range $r_c$ of an active link.

%%%%%%%%%%%%%%%%%%%%%%%%
\subsection{Objectives}

The objective of link scheduling is to maximize the spatial channel-reuse at a time instance, while satisfying the SINR requirement of active links, i.e., 
\begin{eqnarray}
\label{objective}
\mbox{maximize}   & & \sum_{ij} (R_0-\mbox{R}_{ij}) \sigma_{ij} \\
\mbox{subject to} & & \mbox{R}_{th} \ge \mbox{R}_{ij}, \; \mbox{for} \; \forall \; \sigma_{ij}=1, \nonumber
\\ \nonumber
\end{eqnarray}
\noindent
where $\mbox{R}_{ij}$ is the inverse of the interference at receiver $j$ from (\ref{R_gamma}). $R_{th}$ is the inverse of an interference threshold. 
$R_0$ is a large positive constant where $R_0> \mbox{R}_{ij}$ for $\forall$ ($i,j$) $\in$ $\mbox{\boldmath$L$}$. $R_0$ is used for the objective function to be maximized 
whenever a new link becomes active.
The first term $ \sum_{ij} \sigma_{ij}$ is the total number of active links which should be maximized for the channel reuse in the network. The second term  $\sum_{ij} \mbox{R}_{ij} \sigma_{ij}$ represents the effect of total interference from active links. 

The objective function in (\ref{objective}) satisfies the following three properties. The objective function is $(i)$ constrained to satisfy the SINR requirement of any active links; $(ii)$  monotonically increasing with respect to the number of active links;
and
$(iii)$ monotonically increasing as the inverse SINR of an active link decreases. Therefore, the two terms in the objective function make trade-offs between the channel reuse and the SINR requirements.

To measure the optimality of the objective function, we use the link outage probability $P(\mbox{SINR}_{ij} < \mbox{SINR}_{th})$. As soon to be shown, a limited neighborhood size is a major cause of significant link outage probability considered in this work. 

%%%%%%%%%%%%%%%%%%%%%%%%%%%%%%%%%%%%
\subsection{Probabilistic Decisions}
\label{ProbDecision} 

Optimization  is achieved through distributed decisions at individual nodes. Nodes determine whether and when $\sigma_{ij}$'s become active. Decisions made by a node need to account for loss variables, i.e., the aggregated information outside its neighborhood, and decisions made by neighbors. As deterministic distributed decisions can be easily trapped in local minima,  we consider probabilistic, i.e., randomized decisions  (see \cite{Geman} \cite{Bruce} \cite{IEEESigProc} and references therein). At each time epoch, nodes make the link-scheduling decisions based on a probabilistic model of the neighboring nodes' link-scheduling decisions at the previous time epoch and the loss variable. This requires a probabilistic decision model.

%%%%%%%%%%%%%%%%%%
\section{Probabilistic Model}
\label{WNModel}

The probabilistic model of distributed link-scheduling decisions is based on a machine learning framework in form of a Boltzmann distribution and a dependency graph. The model is drawn from our prior work \cite{IEEESigProc}, and included here for completeness.

%%%%%%%%%%%%%%%%%%%%%%%%%%%%%%%%%
\subsection{Boltzmann Distribution}
\label{Boltz}

Our model considers $\sigma_{ij}$'s as a random field of link activities.  Such a random field resembles a particle system with interacting particles (spins) \cite{Yedidia}. States of particles can be viewed as binary random variables. A configuration $\omega$ of interacting binary-state particles can be described by a system potential energy $H(\omega)$. The probability distribution of $H(\omega)$ obeys the Boltzmann distribution $P(\omega)$ = $Z_0^{-1}
\cdot$ $\mbox{exp} \left (-H(\omega) \right)$ \cite{Geman}\cite{Yedidia}, where $Z_0=\sum_{ \omega} \exp \left (-H(\omega)\right )$ is a normalizing constant.

We now regard the objective (\ref{objective}) as the system potential energy and appeal to Boltzmann law in statistical physics \cite{Geman}\cite{IEEESigProc}\cite{Liu}\cite{Tan}\cite{Yedidia}. 
In particular,   the system potential energy of binary link-scheduling decisions is defined as, 
{\small
\begin{eqnarray}
\label{Power4_1}
H(\mbox{\boldmath$\sigma$})
&=& \sum_{ij} (-R_0+\mbox{R}_{ij})\sigma_{ij} + \beta \sum_{ij} U(\mbox{R}_{ij} - \mbox{R}_{th})
\sigma_{ij}  \\
&=& \sum_{ij} (-R_0+{N_b \over P_i l_{ij}^{-\alpha}} ) \sigma_{ij} \nonumber \\
& & + \sum_{ij} \sum_{mn \in N_{ij}} { P_m l_{mj}^{-\alpha}  \over P_i l_{ij}^{-\alpha}} \sigma_{ij} \sigma_{mn} \nonumber \\
& & + \sum_{ij} {h_{ij} \over P_i l_{ij}^{-\alpha}} \sigma_{ij}  + \beta \sum_{ij} U(\mbox{R}_{ij} - \mbox{R}_{th}) \sigma_{ij}, \nonumber
\end{eqnarray}
}
\noindent
where $\beta > 0$ is a constant. $\beta  U(\mbox{R}_{ij} - \mbox{R}_{th})$ is  a penalty term for the SINR constraint. $N_{ij}$ denotes the set of links within the neighborhood range from  receiver $j$ of link ($i,j$), and $h_{ij}$ denotes total residual interference outside the neighborhood range experienced by active link ($i,j$), 
i.e., $h_{ij}$ = $\sum_{mn \not\in N_{ij}}$ $P_m l_{mj}^{-\alpha}$$\sigma_{mn}$.

From Boltzmann law \cite{Geman}\cite{IEEESigProc}, the probabilistic model for a set of link-scheduling decisions at a time instance is, 
\begin{equation}
\label{Eq5}
P(\mbox{\boldmath$\sigma$}) =  Z_{\sigma}^{-1} \cdot \mbox{exp} \left ( {-H(\mbox{\boldmath$\sigma$}) }
\right ),
\end{equation}
\noindent
where $\mbox{\boldmath{$\sigma$}}$ is a random field (i.e., configuration) of all link activities $\sigma_{ij}$'s. 
$Z_{\sigma}$ = $\sum_{\mbox{\boldmath$\sigma$}}$ 
$\mbox{exp} \left (-H(\mbox{\boldmath$\sigma$})\right )$ is a normalizing constant. Note that both the number of active links and interference play important roles: If multiple configurations exist with the same number of active links, the system potential energy distinguishes   different configurations by the sum of the inverse SINR of active links.

A desirable configuration corresponds to a small system potential energy, and a large Boltzmann probability \cite{Geman}. 
Hence, an optimal decision ${\mbox{\boldmath$\sigma$}}^*$ is the one that maximizes the Boltzmann distribution, 
\begin{eqnarray}
{\mbox{\boldmath$\sigma$}}^*
&=& \mbox{arg} \mathop{\mbox{max}}_{\tiny \mbox{\boldmath$\sigma$}} P(\mbox{\boldmath$\sigma_{}$}).
\end{eqnarray}

%%%%%%%%%%%%%%%%%%%%%%%%%%%%%%%%%%%%%%%%%%%%%%%%%%%%%%%%%%
\subsection {Dependency Graph}

\begin{figure}[htb!]
\epsfysize=1.2in
\centerline{\epsffile{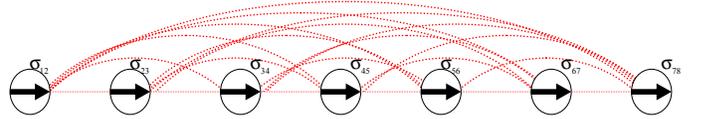}}
\caption{Dependence Graph of random field $\mbox{\boldmath$\sigma$}$}
\label{DSigma}
\end{figure}

The probability model exhibits a graphical representation \cite{IEEESigProc}\cite{Jordan}\cite{Kschischang}. The graphical representation shows the statistical spatial dependence among link activities. For ease of illustration, consider a directional linear topology of eight nodes and seven directional links, where the directions correspond to a data flow. The corresponding  dependency graph of link-scheduling decisions $\mbox{\boldmath$\sigma$}$ is  shown in Figure \ref{DSigma}. 

\begin{figure}[htb!]
\epsfysize=1.5in
\centerline{\epsffile{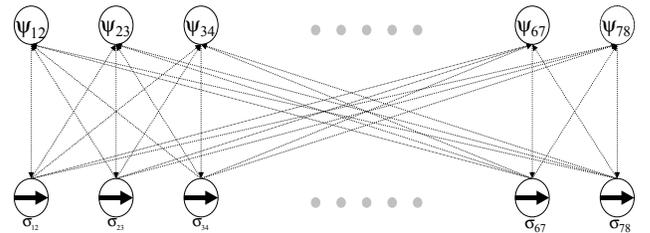}}
\caption{Example Factor Graph Representation.}
\label{global_Factor}
\end{figure}

Nodes in the graph represent binary random variable \{$\sigma_{ij}$\} for $\sigma_{ij}$ $\in$ $\{0,1\}$ and $1 \le i \le 7$ and $j=i+1$. A link between two nodes  represents the spatial dependence. The spatial dependence results from interference in the wireless channel and corresponds to coupled terms in the system potential energy $H(\mbox{\boldmath$\sigma$})$. The dependency graph can be further represented in form of a factor graph  shown in Figure \ref{global_Factor}. A factor graph consists of variable nodes and functional nodes \cite{Kschischang}. Here, $\sigma_{ij}$ denotes a variable node for link ($i,j$) $\in$ $\mbox{\boldmath$L$}$. A functional node $\psi_{{}_{ij}}$ = $R_{ij}$ denotes a potential function of $\sigma_{ij}$.


Link-scheduling decisions are all dependent due to interference. Thus the dependency graph is fully connected. Such a fully connected  graph shows that the neighborhood of a node is in fact the entire network. 

%%%%%%%%%%%%%%%%%%%%%%%%%%%%%%%%%%%%%%%%%%%%%%%%%%%%%%%%%
\subsection{Optimal Configuration and Distributed Decisions}
\label{Opt}

A fully connected dependency graph impacts how distributed decisions should be made using the probabilistic model.   Node decisions are all dependent. Hence, optimal decisions require nodes to exchange information with all the other nodes in the network. Graphically, this means that an entire network is a neighborhood. That is, a node needs to obtain information on all other nodes. This, in reality, prohibits implementation of optimal distributed link-scheduling.

%%%%%%%%%%%%%%%%%%

\section{Approximation of Probabilistic Model}
\label{Graph3}

We now derive approximations of the probabilistic model for distributed scheduling. Our approximation considers a moderate neighborhood for each node where the node has complete information on its neighbors.  Outside the neighborhood, the residual interference, instead of being ignored, is characterized as a random loss variable. Either the mean values or the distributions of the loss variables are then included in the probabilistic model. The resulting model $P^l(\mbox{\boldmath$\sigma_{}$})$ is an approximation of  the exact model $P(\mbox{\boldmath$\sigma_{}$})$.

%%%%%%%%%%%%%%%%%%%%%%%%%%%%%%%%%
\subsection{ Approximation by Mean Field } 
\label{MF}

We first characterize the residual interference by Mean Field (MF) approximation. The MF approximation optimizes a variational bound between the approximated and the exact models\cite{MacKay}, resulting in a set of deterministic mean-field equations. The "mean-values" of the loss variables can then be obtained through solving the Mean-Field equations iteratively. 
To derive the Mean-Field equations, we consider the system potential energy in (\ref{Power4_1}). For simplicity, we assume that  the SINR constraint is satisfied so that the system potential energy is reduced to
{\small
\begin{eqnarray}
H(\mbox{\boldmath$\sigma$}) 
&=& \sum_{ij}f_{ij}(\mbox{\boldmath$\sigma$})  \\ 
&=& \sum_{ij} a_{ij} \sigma_{ij} + \sum_{ij}\sum_{mn \in N_{ij}} a_{ij,mn} \sigma_{ij}\sigma_{mn}  \nonumber \\ \nonumber
& & + \sum_{ij} h_{ij}, \nonumber
\end{eqnarray}
}
\nonumber
where $a_{ij}$ = $-R_0+{N_b \over Pl_{ij}^{-\alpha}}$, $a_{ij,mn}$ = ${P_ml_{mj}^{-\alpha} \over P_i l_{ij}^{-\alpha}}$ from (\ref{Power4_1}), and $h_{ij}=\sum_{mn \not\in N_{ij}} a_{ij,mn} \sigma_{ij}\sigma_{mn}$ corresponds to the aggregated residual interference outside the neighborhood for link $(i,j)$. Note that this  system potential energy  represents a second-order Ising model in statistical physics, where all terms are dependent. 

Let $h_{ij}^*$ be a deterministic approximation of $h_{ij}$. An approximated system potential energy absorbs the impact of the aggregated residual interference into the coefficient $a_{ij}$'s, 
{\small
\begin{eqnarray}
\label{MF_SystemPotentialFunction}
H^l(\mbox{\boldmath$\sigma$})  
&=& \sum_{ij} g_{ij}(\mbox{\boldmath$\sigma$}) \\ 
&=& \sum_{ij} b_{ij} \sigma_{ij} + \sum_{ij}\sum_{mn \in N_{ij}} a_{ij,mn} \sigma_{ij}\sigma_{mn}, \nonumber
\end{eqnarray}
}
where $b_{ij}$ = $a_{ij} + h_{ij}^*$, and $ h_{ij}^*$ is yet to be found. 

We denote the true free energy \cite{MacKay} with $F$, 
\begin{equation}
\beta F = -\ln(Z_P),
\end{equation}
\noindent
where $\beta>0$ is a constant, and $Z_P$ = $\sum_{\mbox{\boldmath$\sigma$}} \exp(-H(\mbox{\boldmath$\sigma$}))$. Let variational free energy \cite{MacKay} be $\tilde{F}$, 
\begin{equation}
\label{PartialZQoverZQ}
\beta \tilde{F} = -\ln(Z_Q),
\end{equation}
\noindent
where $\beta$ is a large positive constant, and $Z_Q$ = $\sum_{\mbox{\boldmath$\sigma$}} \exp(-H^l(\mbox{\boldmath$\sigma$}))$. 

\indent
To find the simplest, i.e., the first-order Mean Field (MF) approximation, we make the following assumptions: ($a$) for link ($i,j$) $\neq$ ($m,n$), the impacts only affect the first order terms; ($b$) for ($m,n$) $\not\in$ $N_{ij}$, $\sigma_{ij}$ is independent of $\sigma_{mn}$; and ($c$) for the approximated system potential energy, the corresponding distribution is $Q(\mbox{\boldmath$\sigma$})$ = $Z_Q^{-1} \exp(-H^l(\mbox{\boldmath$\sigma$}))$. The mutual information measure between the approximated and the true distribution results in an objective function, i.e., a variational free energy \cite{MacKay}, 
{\small
\begin{eqnarray}
\label{PartialVE}
\beta \tilde{F} &=& E_Q(\ln({Q(\mbox{\boldmath$\sigma$}) \over P(\mbox{\boldmath$\sigma$})})) - \ln(Z_P) \\ \nonumber
                &=& E_Q(\ln(H(\mbox{\boldmath$\sigma$}))-\ln(H^l(\mbox{\boldmath$\sigma$})))-\ln(Z_Q) \\ \nonumber
                &=& -\ln(Z_Q)+\sum_{ij}(-h_{ij}\mu_{ij}) 
                    + \sum_{ij} \sum_{mn \not \in N_{ij}} a_{ij,mn} \mu_{ij}\mu_{mn}, \nonumber
\end{eqnarray}
}
\noindent
where $\mu_{ij}$=$E(\sigma_{ij})$=$P(\sigma_{ij}=1)$. Inserting Equation (\ref{PartialZQoverZQ}) in (\ref{PartialVE}) and letting 
${\partial \tilde{F} \over \partial h_{ij}}$=0, 
the optimal value of $h_{ij}$ satisfies 
\begin{equation}
\label{h}
h_{ij}^{*} = \sum_{mn \not \in N_{ij}} a_{ij,mn} \mu_{mn}.
\end{equation}
\indent
$h_{ij}^{*}$ is a weighted sum of expected values of link activities outside the neighborhood range. Hence, $h_{ij}^{*}$ can be regarded as the "mean field" outside the neighborhood of link $(i,j)$. 
As a special case when $a_{ij,mn}=0$ for $mn \not \in N_{ij}$, $h_{ij}^{*}=0$. 

For ($i,j$) $\in L$, there are $n$ such equations, where $n$ is the total number of nodes in a network. Those equations are coupled. The coupling can be shown explicitly by rewriting $h_{ij}^*$ as 
$h_{ij}^*$ = $\log {1-\mu_{ij} \over \mu_{ij} }  {\eta_{1} \over \eta_{2}}$. 
Equation (\ref{h}) then takes the following form,
\begin{equation}
\label{hc}
\log {1-\mu_{ij} \over \mu_{ij}}  {\eta_{1} \over \eta_{2}} 
= \sum_{mn \not \in N_{ij}} a_{ij,mn} \mu_{mn}, 
\end{equation}
where $\eta_{1}$ and $\eta_{2}$ are functions of $h_{kl}^*$'s for 
$(k,l) \ne (i,j)$ (and in turn $\mu_{kl}$'s). 
Hence, the set of $n$ equations are coupled through unknown values of $h_{ij}^*$'s or $\mu_{ij}$'s. Appendix \ref{Derivationhc} provides detailed  derivations. 

The coupled Mean-Field equations can be solved iteratively and numerically starting from an initial condition \cite{MacKay}. The Mean-Field entity $h_{ij}^*$ can also be estimated when measurements are available. For example, the interference within the neighborhood range can be known with the information exchange. 
Thus, if we assume that the aggregated interference can be measured possibly with a measurement error, $h_{ij}^{*}$ can be estimated  when  interference measurements are available.

An advantage of the Mean-Field approximation is that a resulting probability model is a simple Markov random field, where the deterministic Mean-Field values $h*_{ij}$'s absorb the long-range spatial dependence in the residual interference. However, the Mean-Field approximation exhibits two disadvantages. First,  it can be computationally intensive to solve a set of  Mean-Field equations. Second,  if the Mean-Field parameters are  estimated by measurements instead, such measurements on the aggregative residual interference are either unavailable or computationally intensive to obtain. For example, such measurements require nodes to transmit test signals during link-scheduling decisions. In addition, nodes need to be capable of measuring the aggregated  interference. For example, packet reception ratio (PRR) versus SINR models need to be seeded by $O(n)$ trials in an n-node network where each node transmits  while receivers measure the channel conditions \cite{SLiu}.

%%%%%%%%%%%%%%%%%%%%%%%%%%%%%%%%%
\subsection{ Approximation by  Lyapunov Central Limit Theorem} 
\label{ResidualInterference}

An alternative approach is to treat the aggregated residual interference as random and approximate its distribution. Such an approximation can be computationally less expensive to obtain than the Mean-Field approximation. 

Let $n_{ij}$ be a random variable that approximates the residual interference of an active link ($i,j$).
Our goal is to identify a probability distribution for $n_{ij}$. As residual interference aggregates interference signals from many interferers, one option is to approximate $n_{ij}$ as a normal random variables. 

We first obtain an expression for $n_{ij}$ to understand  whether a normal  approximation is feasible.  Based on our assumptions, interferers have the same transmission power $P_0$. Locations of interferers outside a neighborhood range can be arbitrary/random but follow certain patterns of node-positions within the neighborhood. In other words, the network topology exhibits a certain regularity. Specifically, the following assumptions are posed on the location of interferers.

\begin{itemize}
\item{ For an active link ($i,j$), all interferers are located uniformly within radial bands bounded by concentric  circles from receiver $j$. The radius of the circles exhibits randomness, i.e.,  $r_k$ = $r_{k-1}+x_k$ for $k \ge 1$, where $r_0$=$\gamma_f$, and $x_k$=$x$ is a uniform random variable between $r^l$ and $r^u$. Here, $r^l$ and $r^u$ are the minimum and maximum distance between any two active links, within the neighborhood of link ($i,j$). The circle with radius $r_k$ is called the $k$-th frontier. 
}

\item{On the $k$-th frontier for $k \ge 1$, any two neighboring active links are separated by random distance $x_0$. Random variables $x_k$ and $x_0$ are independent and identically distributed.}
\end{itemize}

\begin{figure}[htb!]
\epsfysize=1.5in
\centerline{\epsffile{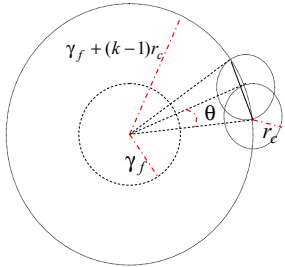}}
\caption{Configuration of Active Links outside Neighborhood range. }
\label{Cardinality_dfactor}
\end{figure}

\indent
Lemma 1: With the above two assumptions, the residual interference $h_{ij}$ for active link ($i,j$) can be modeled as  
\begin{equation}
\label{nij}
n_{ij}= \sum_{k=1}^{\infty} (P_0 r_k^{-\alpha}) \cdot (2 \pi r_k/x_0),
\end{equation}
\noindent
where $r_k$ = $r_{k-1}+x_k$ with $r_0$=$\gamma_f$ as a constant. $x_k=x$ is a random variable for $k \ge 1$. $x$ and $x_0$ are both independent and identically distributed with a uniform distribution between $r^l$ and $r^u$, for $k \ne m$.\\
\indent
The proof can be found in Appendix \ref{Lemma11}. This lemma shows that random positions of interferers outside a neighborhood follow similar characteristics to within the neighborhood.\\

Next, we derive the mean and the variance of $n_{ij}$'s.  
For  analytical feasibility of obtaining the mean value, we approximate the joint probability density 
$P(r_1, \cdots, r_{\infty})$ with a product form based on the Mean-field approximation  \cite{Kappen}.

\indent
Lemma 2: Consider $\{r_1, \cdots, r_{\infty}\}$ as a random field.
Based on the first-order Mean-field approximation, the joint probability density function 
$P(r_1, \cdots, r_{\infty})$ can be approximated as a product of marginal distributions, 
\begin{equation}
\prod_{i=1}^{\infty} P(r_i|<r_{i-1}>),
\end{equation}
\noindent
where $<r_{i-1}>$ = $\gamma_f+(i-1) \overline{r}_c$ denotes the expected value of $r_{i-1}$. $\overline{r}_c$ is an expected value of $r_c$, i.e., $\overline{r}_c$ = $E(r_c)$ = ${r^l+r^u \over 2}$, and $<r_0>$=$\gamma_f$.\\
\indent
The proof can be found in Appendix \ref{Lemma12}. Using such a mean-field distribution, we obtain the expected value  

{\small
\begin{eqnarray}
\label{enij}
E(n_{ij}) &=& \int_{r_1,\cdots,r_{\infty}} \int_{x_0} \sum_{k=1}^{\infty} (P_0 r_k^{-\alpha}) \cdot {2 \pi r_k \over x_0} \cdot  \\
          & &  P(r_k|<r_{k-1}>) P(x_0) d_{r_1} \cdots d_{r_{\infty}}d_{x_0}, \nonumber
\end{eqnarray}
}
where $P(r_1|<r_0>)$=$P(r_1|\gamma_f)$. Closed-form expressions can be derived in the following lemma for the individual terms in the sum.

\indent
Lemma 3: Let $u_k$ be the term in the summation in (\ref{enij}), 
where $u_k$ = $2\pi P_0 r_k^{1-\alpha}$ = $2\pi P_0 (r_{k-1}+x_k)^{1-\alpha}$.

Denote the mean and the variance of $u_k$ as $E(u_k)$ and $V(u_k)$ respectively. For $\alpha >2$,
\begin{eqnarray}
E(u_k) &=& {2\pi P_0 (\mathcal{D}_1^{2-\alpha}-\mathcal{D}_2^{2-\alpha}) \over (2-\alpha)(r^u-r^l)},
\nonumber \\ \nonumber
V(u_k) &=& E(u_k^2)-E(u_k)^2, \nonumber
\end{eqnarray}
where $\mathcal{D}_1$ = $\gamma_f+(k-1)\overline{r}_c+r^u$, $\mathcal{D}_2$ = $\gamma_f+(k-1)\overline{r}_c+r^l$, and
$E(u_k^2)$ $=$ $4\pi^2 P_0^2$
$(\mathcal{D}_1^{3-2\alpha}-\mathcal{D}_2^{3-2\alpha})/((3-2\alpha)(r^u-r^l))$.

Let the $k$-th term of $n_{ij}$ be $v_k$ = ${u_k \over x_0}$. Then,
$E(v_k) = {E(u_k) \ln(r^u/r^l) \over r^u-r^l}$, and 
$V(v_k) = {V(u_k) \over r^u r^l}$.\\
\indent
The proof is given in Appendix \ref{Lemma14}. Using the results from Lemmas 1-3, we obtain the mean and the variance of the residual interferences below. 

\indent

Theorem 1: For a link ($i,j$),  residual interference $n_{ij}$ is a finite sum of independent random variables
with different probability distributions. Thus, the probability distribution of $n_{ij}$ is approximately
normal with mean $E(n_{ij})$ and variance $V(n_{ij})$, 
where $E(n_{ij}) = \sum_{k=1}^{\infty} E(v_k)$, and $V(n_{ij}) =\sum_{k=1}^{\infty} V(v_k)$.\\
\indent
The proof is provided in Appendix \ref{Theorem1}.\\

How accurate is the normal approximation? We check the modeling accuracy using Lyapunov condition. For $n_{ij}$ to converge to a normal distribution, the Lyapunov condition on $n_{ij}=\sum_{k=1}^{\infty}v_k$ should satisfy
{\small
\begin{equation}
{\left (\sum_{k=1}^{\infty} E(|v_k-E(v_k)|^3)\right)^{1/3} \over \left (\sum_{k=1}^{\infty} V(v_k)
\right)^{1/2}} \rightarrow 0.
\end{equation}
}

\begin{figure}[htb!]
\epsfysize=2.2in
\centerline{\epsffile{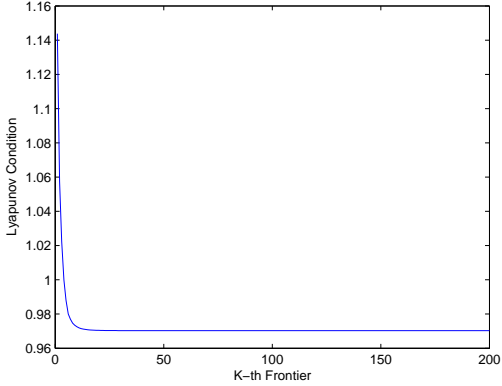}}
\caption{Lyapunov Condition of the residual interference $n_{ij}$}
\label{Lyapunov_Condition}
\end{figure}

\noindent
Such condition is valid for an infinite sum of independent random variables. As the interference from far apart interferers is diminishing to zero, $n_{ij}$ is in reality a sum of a finite number of independent random variables. As a result, the Lyapunov condition converges to a small constant instead of zero. 
A numerical analysis on the normal approximation is conducted on a network with infinite nodes, the increase of neighborhood size to $k$-th frontier neighbors, and channel attenuation parameter $\alpha=4$. 
Figure \ref{Lyapunov_Condition} shows that the Lyapunov condition converges to a small constant around 0.97. 

Hence, the advantage of the normal approximation is the simplicity. Only the mean and the variance need to be estimated. However, the approximation can deviate somewhat from a normal random variable measured by the Lyapunov condition. \\

%

%%%%%%%%%%%%%%%%%%
\subsection{Approximated  Potential Energy and Probabilistic Model}
\label{Sufficient} 

\begin{figure}[htb!]
\epsfysize=2.2in
\centerline{\epsffile{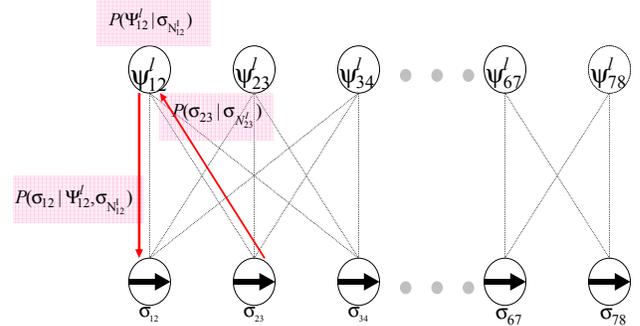}}
\caption{Example Factor Graph with Functional Nodes and Variable Nodes.}
\label{local_Factor}
\end{figure}

The Mean-Field and the normal approximations result in approximated two network potential energy functions and thus the probabilistic models respectively.  First, the Mean-Field approximation in Section \ref{MF} characterizes the residual interference as a deterministic quantity outside the neighborhood. The resulting system potential energy is 
\begin{eqnarray}
H^l(\mbox{\boldmath$\sigma$})  
&=& \sum_{ij} g_{ij}(\mbox{\boldmath$\sigma$}) \\ \nonumber
&=& \sum_{ij} (a_{ij}+h^*_{ij}) \sigma_{ij} + \sum_{ij}\sum_{mn \in N_{ij}} a_{ij,mn} \sigma_{ij}\sigma_{mn}, 
\end{eqnarray}
where $h_{ij}^*$ is deterministic and obtained from the mean-field equation.  The corresponding Boltzmann distribution is
\begin{eqnarray}
\label{pl1}
P^l(\mbox{\boldmath$\sigma$}) &=& Z_{\sigma}^{-1} \mbox{exp} 
\left ( {-H^{l}(\mbox{\boldmath$\sigma$}) } \right ).
\end{eqnarray}
\noindent
Such a Boltzmann distribution is a second-order Markov Random Field with deterministic coefficients $a_{ij}$'s, 
$h_{ij}^{*}$'s and $a_{ij,mn}$'s. 
The resulting probabilistic dependency graph is shown 
in Figure \ref{local_Factor}.

\indent
Second, the normal approximation characterizes the residual interference outside the neighborhood as a Gaussian random variable that aggregates contributions from (unseen) interferers. The resulting system potential energy is,  
{\small
\begin{eqnarray}
\label{pl}
H^{l}(\mbox{\boldmath$\sigma$},\mbox{\boldmath$n$})
&=& \sum_{ij} (-R_0+ {N_b \over P_i l_{ij}^{-\alpha}}) \sigma_{ij}  \\ \nonumber
& & + \sum_{ij} \sum_{mn \in N_{ij}} {P_m l_{mj}^{-\alpha}  \over P_i l_{ij}^{-\alpha}} \sigma_{ij}\sigma_{mn}  
    + \sum_{ij} {n_{ij} \over P_i l_{ij}^{-\alpha}}  \sigma_{ij}  \\ \nonumber
& & + \beta \sum_{ij} U(\mbox{R}^l_{ij} - \mbox{R}_{th}) \sigma_{ij}, \nonumber
\end{eqnarray}
}
\noindent
where $\mbox{\boldmath$n$}$ is a random vector including all $n_{ij}$'s. The corresponding Boltzmann distribution is
\begin{eqnarray}
\label{pl2}
P^l(\mbox{\boldmath$\sigma$},\mbox{\boldmath$n$}) 
&=& Z_{\sigma}^{-1} \mbox{exp} \left ( {-H^{l}(\mbox{\boldmath$\sigma$},\mbox{\boldmath$n$}) } \right ).
\end{eqnarray}
\noindent
This approximated model is a random-bond Markov Random Field \cite{IEEESigProc}, where $n_{ij}$'s are random coefficients (bonds) for $\sigma_{ij}$'s. 
%The resulting probabilistic dependency graph is shown in Figure \ref{local_Factor}. 

As  distributed decisions only need to determine the values of $\sigma_{ij}$'s, 
the marginal probability model for $\sigma_{ij}$'s is 
\begin{equation}
P^l(\mbox{\boldmath$\sigma$}) = \int P^l(\mbox{\boldmath$\sigma$},\mbox{\boldmath$n$}) d\mbox{\boldmath$n$}.
\end{equation}
%

%%%%%%%%%%%%%%%%%%
\section{ Message-Passing over Factor Graph}
\label{SelfManagement}

A simplified probabilistic model $P^l(\mbox{\boldmath$\sigma$})$ from either the mean-field or the normal approximation can be used to determine link activities, 
\begin{equation}
\mbox{\boldmath$\hat{\sigma}$}=\mbox{arg} \mathop{\mbox{max}}_{\mbox{\boldmath$\sigma$}}
P^l(\mbox{\boldmath$\sigma$}). 
\end{equation}

Randomized and distributed decisions implement such optimization through message passing over localized factor graphs. The general message passing algorithm is provided in \cite{Kschischang} and the references therein.  The novelty here is to couple message passing with approximation of the residual interference. The corresponding functional nodes and variable nodes are chosen accordingly to  couple the approximations with distributed scheduling.

%

%%%%%%%%%%%%%%%%%%%%%%%%%%%%%%%%%%%%%%%%%
\subsection{Message Passing with Normal Residual Interference}

The local potential function in message passing,  $P(\psi^l_{ij}|\sigma_{N^I_{ij}})$ for link $(i,j)$, exhibits a special form from approximating residual interference as a normal random variable, i.e., 
{\small
\begin{equation}
\label{Parameters1}
P(\psi^l_{ij}|\sigma_{N^I_{ij}}) = N(E(\psi^l_{ij}),V(\psi^l_{ij})),
\end{equation}
}
\noindent
where $\psi^l_{ij}$=$R^l_{ij}$. $N(E(\psi^l_{ij}),V(\psi^l_{ij}))$ denotes a normal distribution with a mean 
{\small

\begin{equation}
E(\psi^l_{ij}) =
{ \sum_{|X_m-X_j| \le \gamma_f} P_0 l_{mj}^{-\alpha} \sigma_{mn} + N_b  \over P_0 l_{ij}^{-\alpha}} 
+ {1 \over P_0 l_{ij}^{-\alpha}} E(n_{ij}), \nonumber
\end{equation}
}
\noindent
and a variance (from (\ref{R_l_gamma}))
{\small
\begin{equation}
V(\psi^l_{ij}) = ({1 \over P_0 l_{ij}^{-\alpha}})^2 V(n_{ij}). \nonumber
\end{equation}
}

\indent
 Message from a function node $\psi^l_{{}_{ij}}$ to a variable node $\sigma_{ij}$ is 
{\small
\begin{equation}
\label{Parameters2}
P(\sigma_{ij}=1|\psi^l_{ij}, \sigma_{N^I_{ij}}) = U(R_{th}-\psi^l_{ij}),
\end{equation}
}
\noindent
where $U(R_{th}-R^l_{ij})$=1, if $R_{th} > R^l_{ij}$; 0, otherwise.\\

\indent
Message from a variable node $\sigma_{ij}$ to a function node $\psi^l_{{}_{mn}}$ is 
{\small
\begin{equation}
\label{Parameters3}
P(\sigma_{ij}=1| \sigma_{N^I_{ij}})
= \left \{
  \begin{array}{ll}
  {1 \over 2} + Q({ R_{th}-E(\psi^l_{ij})\over V(\psi^l_{ij})}), & E(\psi^l_{ij}) \le R_{th} \\
  Q({ E(\psi^l_{ij})-R_{th} \over V(\psi^l_{ij})}), & \mbox{otherwise,}\\
  \end{array}
\right.\ \nonumber
\end{equation}
}
\noindent
where $Q(x)$=$\int_0^x {1 \over \sqrt{2\pi}} e^{-t^2 \over 2} dt$.  
This equation is obtained from an equality of $P(\sigma_{ij}=1| \sigma_{N^I_{ij}})$ = $\int P(\sigma_{ij}=1| \psi_{ij}^l, \sigma_{N^I_{ij}})$
$P(\psi_{ij}^l|\sigma_{N^I_{ij}}) d\psi_{ij}^l$.\\
\indent

Consider an example in Figure \ref{local_Factor}. The second node at the top row determines the scheduling of link ($2,3$). This node maintains only  the local factor graph related to link ($2,3$). The local graph has variables  
$\psi_{23}^l$, $\sigma_{12}$,$\sigma_{23}$ and $\sigma_{34}$. 

The algorithm includes two steps: (a) estimating the parameters for the mean and variance of the residual interference, (b) using the estimated parameters in the local functions for message passing. 
For example, in Figure \ref{local_Factor}, a functional node $\Psi^l_{12}$ estimates the mean and variance of residual interference $n_{12}$, $E(n_{12})$ and $V(n_{12})$, from the received message from neighboring variable nodes $\sigma_{23}$ and $\sigma_{34}$. The estimation is from Theorem 1.

The functional node $\psi^l_{12}$ sends to a variable node $\sigma_{12}$ a message $P(\sigma_{12}=1|\psi_{12}^l, \sigma_{N^I_{12}})$ = $U(R_{th}-\psi_{12}^l)$. 
The variable $\sigma_{12}$ sends to all the neighboring functional nodes, $\Psi_{12}^l$, $\Psi_{23}^l$ and $\Psi_{34}^l$, a message $P(\sigma_{12}=1|\sigma_{N^I_{12}})$ in (\ref{Parameters3}).

%%%%%%%%%%%%%%%%%%%%%%%%%%%%%%%%%%%%%%%%%%%%%%5
\subsection{Message Passing with Mean-Field Residual Interference}

The local function with the Mean-Field approximation of the residual interference, i.e.,  $h_{ij}^*$ in (\ref{MF_SystemPotentialFunction}), is relatively simple, 
{\small
\begin{equation}
P(\psi_{ij}^l=a_0|\sigma_{N^I_{ij}}, h_{ij}^*) = \delta(\psi_{ij}^l-a_0), 
\end{equation}
}
\noindent
where $\psi_{ij}^l$=$R_{ij}^l$ in (\ref{R_l_gamma}) with $\mbox{Res}_{ij}$=$h_{ij}^{*}$ in (\ref{MF_SystemPotentialFunction}). $a_0$ is a constant. $\delta(\psi_{ij}^l-a_0)$=1, if $\psi_{ij}^l = a_0$; 0, otherwise.\\

\indent
Message from a function node $\psi_{{}_{ij}}^l$ to a variable node $\sigma_{ij}$ is
{\small
\begin{equation}
P(\sigma_{ij}=1|\psi_{ij}^l, \sigma_{N^I_{ij}}, h_{ij}^*) = U(R_{th}-\psi_{ij}^l), 
\end{equation}
}
where $U(x)$ is a unit step function.

\indent
Message from a variable node $\sigma_{ij}$ to a function node $\psi_{{}_{mn}}^l$ is
{\small
\begin{equation}
P(\sigma_{ij}=1| \sigma_{N^I_{ij}}, h_{ij}^*) = U(R_{th}-\psi_{ij}^l).
\end{equation}
}

The overall algorithm includes two steps: (a) estimating the mean-field parameter $h_{ij}^*$'s, and (b) obtaining decisions $\sigma_{ij}$'s through message passing. These two steps alternate until convergence.

%%%%%%%%%%%%%%%%%%%%%%%%%%%%%%%
\section{Simulation}
\label{PE}

We now evaluate network-wide link outage probability to assess the benefit of the approximation on link-scheduling. As the link-outage probability is not analytically related yet with the network model, we conduct simulations for the shared channel access on a multi-hop wireless network.

%%%%%%%%%%%%%%%%%%%%%%%%%%%%
\subsection{Simulation Setup}

Network nodes are positioned uniformly in a square area of 100 square meters, and composed of $L$ = 200 links. We consider $2 \le \alpha \le 6$ for the channel attenuation factor. Here, $2 \le \alpha \le 3$ results in 'strong' dependence among interferers. This corresponds to the scenario which is unsolved in our prior work. We choose $10 \le \mbox{SINR}_{th} \le 100$. We consider the maximum spatial channel-reuse at a time instance. 

%%%%%%%%%%%%%%%%%%%%%%%%%%%%%%
\subsection{Simulation Results}

\begin{figure}[htb!]
\epsfysize=3.0in
\centerline{\epsffile{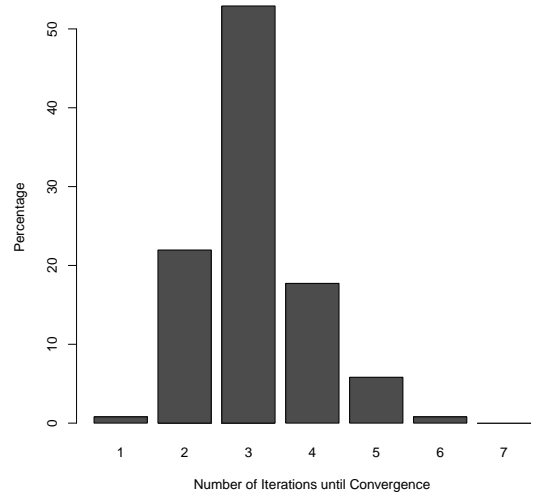}}
\caption{Convergence of Distributed Link-Scheduling Decisions}
\label{IterationsUntilConvergence}
\end{figure}

Using the above experimental setting, we first study through simulation whether distributed decisions converge sufficiently fast. Figure \ref{IterationsUntilConvergence} shows that the distributed link-scheduling decisions converge through interactions among neighbors in a few iterations.

We now compare the performance of the following algorithms: (a) distributed link scheduling with the Mean-Field approximation, (b) distributed link scheduling with the normal approximation, and (c) the conventional distributed link-scheduling algorithm \cite{Bjorklund}\cite{Gandham}\cite{IEEESigProc} that simply ignores the residual interference outside neighborhood range. The link outage probability is used as a performance measure.  Such a performance measure assesses  the probability that a link cannot satisfy  the SINR requirement. Here, the link outage is  mainly caused by the limited neighborhood size. There, improved performance signifies the importance of proper approximations of information outside a neighborhood. 

\begin{figure}[htb!]
\epsfysize=3.0in
\centerline{\epsffile{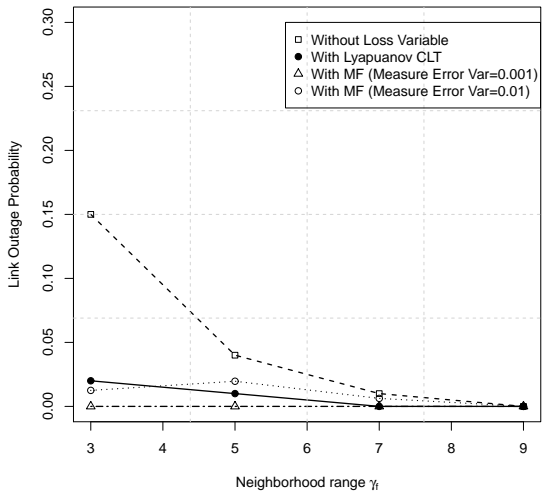}}
\caption{Link Outage probability with $\mbox{SINR}_{th}$=10. Dotted  line: Conventional scheme. Other curves: Our approximation approaches}
\label{SuccessRatioWrtRf}
\end{figure}

Figure \ref{SuccessRatioWrtRf} plots the outage probability as a function of neighborhood size $\gamma_f$ assuming that all nodes have the same size neighborhood. This figure shows that our two approximations for the information outside neighborhoods result in a small link outage probability given for a small neighborhood range $\gamma_f$.  The conventional scheme, however, has a large link-outage probability for a small $\gamma_f$. This shows that  for a small neighborhood, learning the residual interference with a loss variable provides a big gain. The performance with the mean-field  approximation varies with respect to the measurement error of residual interference. When the measurement error is small, the Mean-Field approximation results in a small link-outage probability across different neighborhood sizes. Overall, the two approximation approaches outperform significantly the method that ignores the residual interference, showing the importance of approximating the information outside a neighborhood.

\begin{figure}[htb!]
\epsfysize=3.0in
\centerline{\epsffile{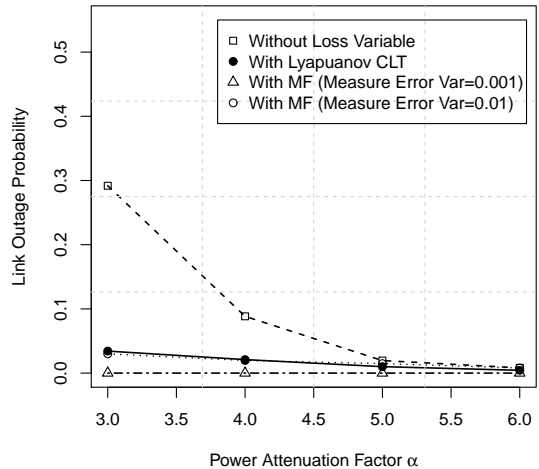}}
\caption{Link Outage probability with $\gamma_f$=4 and $\mbox{SINR}_{th}$=10. Dotted line: The conventional scheme. Other curves: The two approximation schemes.}
\label{SuccessRatioWrtAlpha}
\end{figure}

Figure \ref{SuccessRatioWrtAlpha} shows  how the link-outage probability varies with respect to channel attenuation factor $\alpha$. The link outage probability for the two approximation schemes remains small for different $\alpha$. As $\alpha$ increases, the residual interference outside the neighborhood decrease, and the  link outage probability of the two approximation schemes approaches  zero. It is worth mentioning that for small $\alpha$, e.g., $\alpha=3$, the two approximation approaches significantly outperform the conventional approach. 
The scenario of small $\alpha$ is shown in our prior work \cite{IEEESigProc} to be difficult for distributed scheduling with information only from neighbors. Hence, the result here shows the importance and the ability of the approximations to account for the aggregated residual interference outside a neighborhood. 

\begin{figure}[htb!]
\epsfysize=3.0in
\centerline{\epsffile{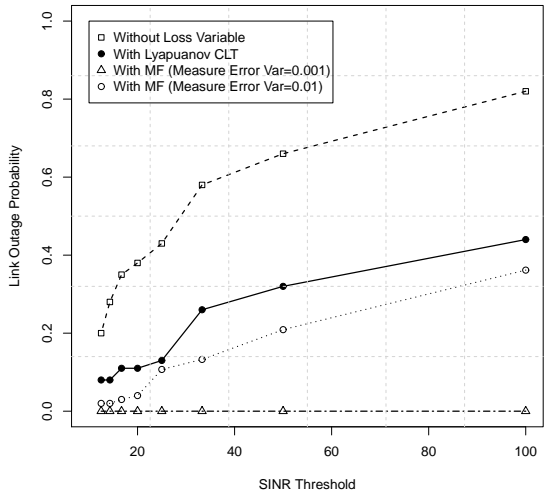}}
\caption{Link Outage probability with $\gamma_f$=3. Dotted line:  Conventional scheme. Other curves: Approximation schemes.}
\label{SuccessRatioWrtSINRthWithRf3}
\end{figure}

\begin{figure}[htb!]
\epsfysize=3.0in
\centerline{\epsffile{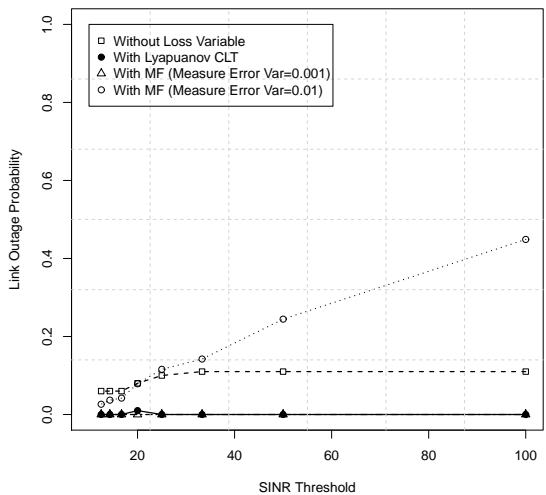}}
\caption{Link Outage probability with $\gamma_f$=4.  Dotted line:  Conventional scheme. Other curves: Approximation schemes. }
\label{SuccessRatioWrtSINRthWithRf4}
\end{figure}

Figure \ref{SuccessRatioWrtSINRthWithRf3} and \ref{SuccessRatioWrtSINRthWithRf4} show the performance as a function of  SINR.  As SINR threshold increases, the link outage probability increases proportionally except for the Mean-Field approximation with small measurement error. The performance of the Mean-Field  approximation for a large measurement error is worse as the SINR threshold increases. This is because the SINR constraint  is more easily violated due to the measurement error as the SINR threshold increases.

For the approximation scheme with Lyapunov central limit theorem (CLT), once $\gamma_f$ is moderately large, the proposed scheme copes well with a large SINR threshold. When neighborhood size  $\gamma_f$ is not sufficiently large, for a moderate  SINR threshold, conventional schemes in the literature often fail to satisfy the SINR requirement, resulting in  a high link outage probability.  On the contrary, across a wide range of SINR threshold values, the proposed scheme with Lyapunov CLT performs  efficiently and improves the link outage probability significantly. In particular, if $\gamma_f$ is moderate (e.g., $\gamma_f=4$ in Figure \ref{SuccessRatioWrtSINRthWithRf4}), the link outage probability quickly approaches zero. Therefore, with both approximation approaches, resource utilization is significantly improved  overall.

%%%%%%%%%%%%%%%%%%%%%%%%%%%%%%%%%%%%%%%%
\section{Conclusion}
\label{Conclusions}

We have studied in this work distributed scheduling decisions where each wireless node uses complete information within a neighborhood and inferred information outside. We have characterized the complex spatial dependence of distributed link-scheduling decisions with a factor graph. The factor graph enables information exchange among neighbors. More importantly,  the long-range dependence in residual interference is not ignored but approximated by one variable as the aggregated dependence outside a neighborhood. The Mean Field approximation considers the residual interference outside a neighborhood as a deterministic "Mean-Field". Such Mean-Field approximation optimizes the variational free energy, resulting in a simple Markov Random Field model for distributed decisions. A disadvantage is that the Mean-Field estimates can be computationally costly to obtain. The normal approximation considers the residual interference outside a neighborhood as a random loss variable. Such a loss variable is approximately normal by the Lyapunov central limit theorem, resulting in a simple Random Bond model for distributed scheduling decisions. A disadvantage is that the condition for the asymptotic normality can  be somewhat deviated for a network with a finite size. 

The approximated dependencies are easily incorporated into the factor graph. This results in an extended message-passing algorithm with the two phases. First, the approximations are obtained iteratively from the available information at neighboring nodes. Second, the approximations are combined with messages from neighboring nodes and allow each node to make its link-scheduling decisions. The algorithm alternates between the two phases, and has been shown to converge in a few than $10$ iterations. 

We have shown through simulations that the link outage probability of the resulting link-scheduling decisions is significantly improved by the approximated information. The improvement is especially pronounced for distributed decisions with a small number of neighbors and a slowly-varying channel, where the link-outage probability is reduced from, e.g., $0.3$ to $0.04$. 

An open issue as possible future work is to derive an analytical relationship   between the network-wide link outage probability and the approximated information.  

%%%%%%%%%%%%%%%APPENDIX %%%%%%%%%%%%%%%%%%%%%%%%%%%%%%%%%%%%%%%%

\section*{Appendix}
\setcounter{section}{0}

\section{Derivation of Equation (\ref{hc})}
\label{Derivationhc}

$\mu_{ij}=P(\sigma_{ij}=1)$, where the probability is evaluated using the simplified Boltzmann distribution 
$Q(\mbox{\boldmath$\sigma$})$ from the mean-field approximation. Expanding the expression of $\mu_{ij}$, we have
\begin{equation}
\mu_{ij} = {   \exp(-h_{ij}^*) \eta_1  \over \exp(-h_{ij}^*)\eta_1+\eta_2 },
\end{equation}
where $\eta_1$ = $\exp(-a_{ij}) \sum_{(m,n) \ne (i,j)} \exp(-H(\mbox{\boldmath$\sigma$} \setminus \sigma_{ij}) )$ 
with $\mbox{\boldmath$\sigma$} \setminus  \sigma_{ij}$ being all $\sigma_{mn}$'s except for $(m,n)=(i,j)$. 
$\eta_2$ = $\sum_{(m,n) \ne (i,j)} \exp(- H(\mbox{\boldmath$\sigma$} | \sigma_{ij}=0) )$. 

Representing $h^{*}_{ij}$ from the above expression, we have 
\begin{equation}
h_{ij}^*= \log {  1-\mu_{ij} \over \mu_{ij}  }     {\eta_{1} \over \eta_{2}}. 
\end{equation}

Replacing this expression in (\ref{hc}), we have 
\begin{equation}
\log { 1-\mu_{ij} \over \mu_{ij}}   { \eta_{1} \over \eta_{2}}
=\sum_{mn \not \in N_{ij}} a_{ij,mn} \mu_{mn}.
\end{equation}

Note that $\eta_1$ and $\eta_2$ are functions of $h^{*}_{mn}$'s ($\mu_{ij}$ís) for $(m,n) \ne (i,j)$. 
$\mu_{ij}$ is also a function of $h^{*}_{mn}$'s. Hence, the equation is coupled. 
In addition, there are multiple equations that contain common $h_{ij}^{*}$'s. 
Hence, for all $(i,j)$'s, the mean-field equations are coupled, and can be solved iteratively.

\section{Proof of Lemma 1}
\label{Lemma11}

On the $k$-th circle for $k \ge 1$, the total number of active links on the circle is 
${2\pi r_k \over x_0}$, which is random. Each active link on the $k$-th circle results in as much interference as 
$P_0 r_k^{-\alpha}$. As a result, total interference from the active links on the $k$-th circle is 
$(P_0 r_k^{-\alpha}) \cdot {2 \pi r_k \over x_0}$; and, thus 
$n_{ij}$ = $\sum_{k=1}^{\infty}$ $(P_0 r_k^{-\alpha}) \cdot {2 \pi r_k \over x_0}$.

\section{Proof of Lemma 2}
\label{Lemma12}

Mean field approximation shows that the joint distribution of a random field ($y_1, \cdots, y_n$) can be approximated by a product of marginal distributions. Thus, $P(y_1,\cdots, y_n)$ = $\prod_{i=1}^n$
$P(y_i|<y_{N_i}>)$ with a good approximation, where $y_{N_i}$ is the set of neighboring random variables of $y_i$.
From the assumptions, $r_k$=$r_{k-1}+x_k$. Thus, $P(r_k|<r_{N_k}>)$=$P(r_k|<r_{k-1}>)$.

\section{Proof of Lemma 3}
\label{Lemma14}

The average of $u_k$, denoted with $E(u_k)$, can be derived from an integration of 
$\int 2\pi P_0 r_k^{1-\alpha} P(r_k|<r_{k-1}>) dr_k$. The average of $u_k^2$ can be derived in a similar way. 

\section{Proof of Theorem 1}
\label{Theorem1}

Consider the configuration in Figure \ref{Cardinality_dfactor}. For an active link ($i,j$), the residual interference of this configuration is denoted with a random variable $n_{ij}$, which is a function of random variables $r_k$ for $k \ge 1$, where $r_k$ is the radius of the $k$-th frontier.

Note that $r_k$'s are independent random variables with different uniform distributions. Lyapunov's central
limit theorem \cite{Stat} shows that the summation of a large number of  independent random variables (even with
different distribution) results in a normal distribution. $n_{ij}$ is a summation of hundreds and thousands of 
independent random variables (i.e., $r_k$'s), thus $n_{ij}$ can be approximated by a normal distribution. 

Furthermore, a normal distribution can be completely characterized by mean and variance. The mean and variance of $n_{ij}$ is denoted with $E(n_{ij})$ and $V(n_{ij})$, respectively, i.e.,
$E(n_{ij})$ = $\sum_{k=1}^{\infty} E(v_k)$, and
$V(n_{ij})$ = $\sum_{k=1}^{\infty} V(v_k)$.

%%%%%%%%%%%%%%%%%%% REFERENCES %%%%%%%%%%%%%%%%%%%%%%%%%%%%%%%%%%%%%%%%
\nocite{*}
\bibliographystyle{paper}

\end{document}